\documentclass[conference]{IEEEtran}
\IEEEoverridecommandlockouts

\usepackage{cite}
\usepackage{hyperref}
\usepackage{amsmath,amssymb,amsfonts}
\usepackage{algorithmic}
\usepackage{graphicx}
\usepackage{textcomp}
\usepackage{xcolor}
\usepackage{booktabs}
\usepackage{caption}
\def\BibTeX{{\rm B\kern-.05em{\sc i\kern-.025em b}\kern-.08em
    T\kern-.1667em\lower.7ex\hbox{E}\kern-.125emX}}
\begin{document}

\title{Spatio-Temporal Graph Structure Learning for Earthquake Detection\\
\thanks{This work is supported by Tsinghua Shenzhen International Graduate School Start-up fund under Grant QD2022024C, Shenzhen Science and Technology Innovation Commission under Grant JCYJ20220530143002005 and Shenzhen Ubiquitous Data Enabling Key Lab under Grant ZDSYS20220527171406015.}
}

\author{\IEEEauthorblockN{Suchanun Piriyasatit, Ercan Engin Kuruoglu*\thanks{* Corresponding author}}
\IEEEauthorblockA{\textit{Data Science and Information Technology Research Center} \\
\textit{Tsinghua-Berkeley Shenzhen Institute}\\
Shenzhen, China \\
zhangsm22@mails.tsinghua.edu.cn, kuruoglu@sz.tsinghua.edu.cn}
\and
\IEEEauthorblockN{Mehmet Sinan Ozeren}
\IEEEauthorblockA{\textit{Eurasia Earth Sciences Institute} \\
\textit{Istanbul Technical University}\\
Istanbul, Turkey \\
ozerens@itu.edu.tr}

}

\maketitle
\pagestyle{plain}

\begin{abstract}
Earthquake detection is essential for earthquake early warning (EEW) systems. Traditional methods struggle with low signal-to-noise ratios and single-station reliance, limiting their effectiveness. We propose a Spatio-Temporal Graph Convolutional Network (GCN) using Spectral Structure Learning Convolution (Spectral SLC) to model static and dynamic relationships across seismic stations. Our approach processes multi-station waveform data and generates station-specific detection probabilities. Experiments show superior performance over a conventional GCN baseline in terms of true positive rate (TPR) and false positive rate (FPR), highlighting its potential for robust multi-station earthquake detection. The code repository for this study is available at \footnote{\url{https://github.com/SuchanunP/eq_detector}}.
\end{abstract}


\section{Introduction}
Earthquake detection is a task to distinguish earthquake signals from various non-earthquake signals and noise recorded by a seismic sensor. This task relies on a network of seismic stations and plays a critical role in earthquake early warning (EEW) systems \cite{SATRIANO2011106}. Unlike earthquake prediction, which seeks to forecast the time, location, and magnitude of future earthquakes—a goal that remains elusive \cite{1997kagan, 2015radon, radon-2019}—earthquake detection provides an immediate and practical means to mitigate the impacts of earthquakes \cite{10.3389/feart.2021.726045}.

Methods for earthquake detection include the classic STA/LTA thresholding \cite{1998sta-lta}, which is a straightforward calculation of a short-term average over a long-term average, in which a detection is declared when this ratio exceeds a predefined threshold. This technique, despite its simplicity, is not effective to detect low signal-to-noise ratio seismic signals and is susceptible to time-varying background noise.

Other methods include template matching method \cite{Gibbons2006TheDO, brown2008autocorrelation, yoon2015earthquake}, which computes normalized cross-correlations (CC) of an event template waveform with candidate windows. A detection is declared if this value exceeds a predefined threshold. Since a priori waveform template is required, this method has low general applicability.

Given massive seismological data being generated everyday, many researchers are paying more attention to deep learning methods. Recent deep learning approaches \cite{ross2018generalized, perol2018convolutional, mousavi2019bayesian, zhu2019phasenet, 10.1093/gji/ggab401, Zhu_2022, saad-2022} are shown to outperform existing traditional algorithms, however these methods use only single-station waveforms to detect single-station earthquake signals and do not utilize graph structure of a seismic network, despite evidence that waveform data from multiple seismic sensors improves event detection \cite{10.1093/gji/ggy132}.

In this context, graphs provide a powerful means of modeling irregular relationships between multi-station data. Unlike Euclidean-based representations such as grids or matrices, which assume regular, fixed structures, graphs can represent spatial information flexibly, capturing complex and non-uniform relationships between variables. By encoding seismic stations as nodes and inter-station relationships as edges, graph structures enable the incorporation of spatial dependencies that are not restricted to predefined grids. Additionally, edge attributes and graph topology provide a framework for modeling both local and global interdependencies, making graphs well-suited for capturing the relationships inherent in seismic networks.

In deep learning, Graph Neural Networks (GNNs) \cite{ZHOU202057} have emerged as powerful tools for modeling data with complex relationships, where entities (nodes) are interconnected via underlying relationships (edges). Unlike traditional neural networks, which assume a Euclidean structure of data, GNNs operate on graphs, making them well-suited for a wide range of applications, including social networks \cite{li2023survey}, molecular chemistry \cite{wang2023graph}, and traffic prediction \cite{Zhang_Chang_Meng_Xiang_Pan_2020}.

Among the various architectures of GNNs, Graph Convolutional Networks (GCNs), introduced by Kipf and Welling \cite{kipf2017semisupervised}, have gained significant popularity due to their ability to generalize the convolution operation to graph-structured data. This architecture enables GCNs to aggregate node features effectively while incorporating both local and global graph structures. A key subclass of GCNs is spectral GCNs, which are grounded in graph signal processing \cite{shuman2013emerging} and focus on designing spectral graph convolutions in the Laplacian spectral domain \cite{10.1145/3627816}. These methods utilize the eigenvalues and eigenvectors of the Laplacian to define convolutional operations in the graph Fourier domain. This approach allows for the extraction of high-order dependencies in the graph, capturing both global and local structures. To improve computational efficiency, Chebyshev polynomial expansions \cite{NIPS2016_04df4d43} approximate spectral convolutions, reducing the need for explicitly computing eigenvectors.

Related study \cite{kim2021graph} employs a general spatial GCN to classify natural earthquakes, artificial earthquakes, and noise in the Korean region. However, it does not explicitly model temporal dependencies over time windows and focuses primarily on offline classification tasks aimed at categorizing past seismic events, not earthquake event detection. Another study \cite{yano2021graph} integrates graph theory into CNNs to identify earthquake signals within a seismic network but does not incorporate GCNs. Additionally, the model's output is a single detection probability for the entire seismic network in a time window, not a detection time series for each station in a network.

In our work, we adapt the Structure Learning Convolution (SLC) framework \cite{Zhang_Chang_Meng_Xiang_Pan_2020} to incorporate learnable structural information directly into the convolutional process of GCNs for spatial modeling, while employing a Gated Recurrent Unit (GRU) \cite{cho-etal-2014-properties} for temporal modeling. Our approach is designed to detect earthquake signals from multi-station seismic networks, capturing both static and dynamic relationships between stations, rather than relying solely on predefined connections. Unlike a previous method \cite{yano2021graph}, where the model produces a detection time series per a network of seismic stations, our model generates a detection time series for each seismic station in a network, a feature particularly beneficial for earthquake early warning (EEW) systems, as earthquake signals typically reach different stations at varying times \cite{eew-2011}. Moreover, unlike the previous method, our proposed method outputs a detection probability per time step rather than per a time window. This captures the probability variation over time, and is helpful to understand signal progression within the window. Additionally, our method can be adapted for real-time detection by processing data in a sliding window manner.

\section{Waveform Data}
The seismic data used in this study were obtained from the Metropolitan Seismic Observation network (MeSO-net) \cite{sakai2009distribution, aoi2021multi}, a regional seismograph network in the Tokyo metropolitan area, Japan. A portion of the data, covering the period from September 4th to 16th, 2011, was utilized from Yano et al. \cite{yano2021graph}, where P- and S-wave arrival times were visually picked. P-waves, the fastest seismic waves, arrive first and provide initial earthquake information, while the slower S-waves cause more significant ground motion and follow the P-waves. The remaining waveform data covering earthquake waveform data from year 2011-2018, were directly sourced from the National Research Institute for Earth Science and Disaster Resilience (NIED) MeSO-net database \cite{oai:nied-repo.bosai.go.jp:00006038}, and then visually picked P- and S-wave arrival times. The waveform data is in three components (North-South, East-West, and Up-Down). The physical values of the waveform are in $m/s^2$ unit.

To showcase our proposed method, we selected 13 stations located at the eastern part of MeSO-net in the same manner as Yano et al. \cite{yano2021graph}, because these stations are distributed around the center of seismically active area and around Kanto district.

To preprocess waveform data for training and evaluation, for the waveform of each component of each station, we subtracted from  it the mean value (with an exception for the Up-Down component of the data directly sourced from NIED, which stated that the offset has already been subtracted). We then applied a 2-8 Hz bandpass filter to remove background noises \cite{nishida2008three}, and normalize the resulting waveform by dividing by the median of the absolute maximum of the resulting waveform. Finally, we downsampled the waveform from 200 Hz to 25 Hz. 

Only the data from 13th to 16th September, 2011 was used for evaluation. The rest was used for training.

The dataset is labeled with earthquake probabilities, where each label is assigned as follows: a probability of 1 (indicating an earthquake) is assigned from the time of the first P-wave arrival up to a time equal to the first P-wave arrival plus 1.4 times the duration between the S-wave and P-wave arrivals. Beyond this interval, the probability is set to 0. This labeling scheme is designed to capture the key time frame associated with earthquake signal detection.

Due to limited data samples, we augmented data by shifting the training windows and adding noises. The augmentation noises are Gaussian distributed with a zero mean and a standard deviation drawn from an exponential distribution with mean 0.001.

\begin{figure}
    \centering
    \includegraphics[width=0.5\textwidth]{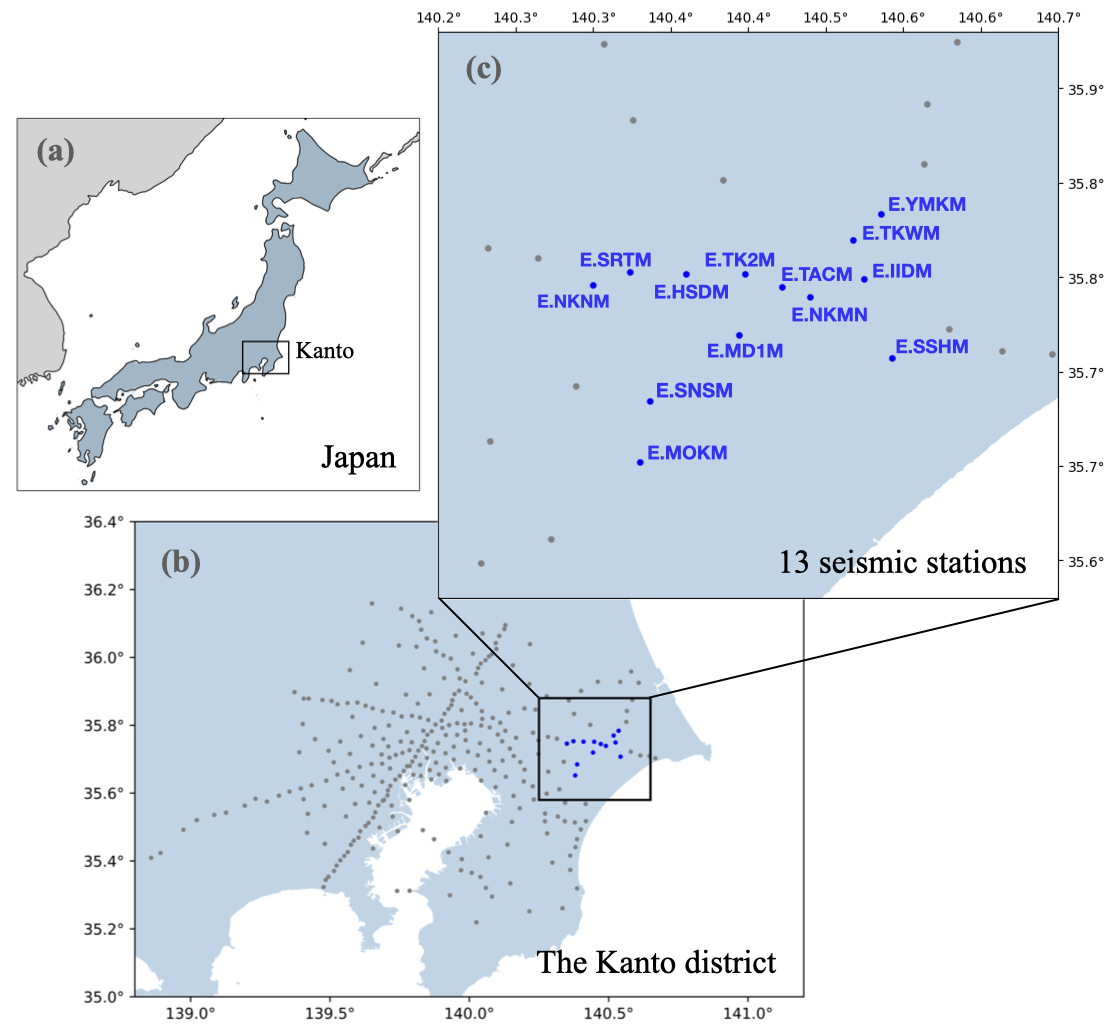} 
    \caption{(a) Map of the Kanto district relative to Japan. (b) Locations of the MeSO-net stations (gray and blue dots) in the Kanto district. (c) The 13 seismic stations (blue dots) used in this study.}
    \label{fig:kanto}
\end{figure}

\section{Methodology}

\subsection{Problem Statement}
In this study, we represent the seismic network as a graph $\mathcal{G} = (\mathcal{V}, \mathcal{E})$, where $\mathcal{V}$ is the set of $N$ seismic stations (nodes), and $\mathcal{E}$ is the set of edges representing the spatial relationships between stations, note that $\mathcal{E}$ is not necessarily given but can be learned. Each node $v_i \in \mathcal{V}$ is associated with a waveform signal $\mathbf{D}_i \in \mathbb{R}^{P \times 3}$, where $P$ is the number of time steps, and $3$ corresponds to the North-South, East-West, and Up-Down components of the waveform data.

The task of earthquake detection is to learn a mapping function $h(\cdot)$, which takes waveform signals $\mathbf{D}$ on a graph $\mathcal{G}$ as input and determines probabilistic values of whether the input waveform signal at each time step was caused by an earthquake event:
\begin{equation}
\tilde{\mathbf{D}}=h(\mathbf{D}, \mathcal{G}, \psi),
\end{equation}
where $\tilde{\mathbf{D}} \in [0, 1]^{P}$ is the detection result and $\psi$ represents the learnable parameters.

\subsection{Spatial Module: Spectral Structure Learning Convolution (Spectral SLC)}
To leverage the spatial relationships in the seismic network, we adapted the spectral component of Structure Learning Convolution proposed in \cite{Zhang_Chang_Meng_Xiang_Pan_2020}, originally used in traffic prediction. The Spectral SLC captures graph structures by combining Chebyshev polynomial-based spectral convolutions and learnable adjacency matrices, which model both static long-term and dynamic input-driven spatial relationships.

Recall that Chebyshev polynomials approximate the graph convolution operation in the spectral domain by indirectly leveraging the eigenvalues of the graph Laplacian, enabling efficient spectral filtering without explicitly computing the eigenvalues or eigenvectors \cite{defferrard2016convolutional}. 

For a given adjacency matrix $\mathbf{A}$ and input feature matrix $\mathbf{X} \in \mathbb{R}^{N \times C_{in}}$ (where $N$ is the number of nodes and $C_{in}$ is the input feature dimension), the $k$-th order Chebyshev polynomial $T_k(\mathbf{A})$ is recursively computed as:
\begin{equation}
T_k(\mathbf{A}) = 2 \mathbf{A} T_{k-1}(\mathbf{A}) - T_{k-2}(\mathbf{A}), 
\end{equation}
with $T_0(\mathbf{A}) = \mathbf{I}$ (identity matrix) and $T_1(\mathbf{A}) = \mathbf{A}$. 

We will explain the adapt the Spectral SLC layer \cite{Zhang_Chang_Meng_Xiang_Pan_2020}, which learns both static and dynamic components of the graph structure. The static component $\mathbf{W}^s$ captures long-term spatial relationships, while the dynamic component $\mathbf{W}^d$ is not a learnable parameter but rather computed dynamically based on the input features as:
\begin{equation}
\mathbf{W}^d = \phi\left(\mathbf{x}, \mathbf{W}_\phi\right)= \mathbf{X}^\top \mathbf{W}_\phi \mathbf{X},
\end{equation}
where $\mathbf{X} \in \mathbb{R}^{N \times C_{i n}}$ is the input of the layer and $\mathbf{W}_\phi$ is a learnable parameter matrix. These components are combined with Chebyshev polynomials to produce the static and dynamic filters:
\begin{equation}
\mathbf{F}_s = \sum_{k=0}^{K-1} \theta_k^s T_k(\mathbf{W}^s),
\quad
\mathbf{F}_d = \sum_{k=0}^{K-1} \theta_k^d T_k(\mathbf{W}^d),
\end{equation}
where $K$ denotes the maximum orders of the Chebyshev polynomials and $\theta_k^s$, $\theta_k^d \in \mathbb{R}^{C_{i n} \times C_{o u t}}$ are learnable parameters.

The output features are obtained by aggregating the static and dynamic filters using a non-linear activation function:
\begin{equation}
\mathbf{H} = \text{ReLU}(\mathbf{F}_s) + \text{ReLU}(\mathbf{F}_d),
\label{eqn:spatial-output}
\end{equation}
where $\mathbf{H} \in \mathbb{R}^{N \times C_{out}}$ is the output feature matrix, and $C_{out}$ is the output feature dimension. Note that the output in Eq. \ref{eqn:spatial-output} represents the learned spatial relationships.

\subsection{Temporal Module: a Gated Recurrent Unit (GRU)}
\begin{figure*}
    \centering
    \includegraphics[width=.7\textwidth]{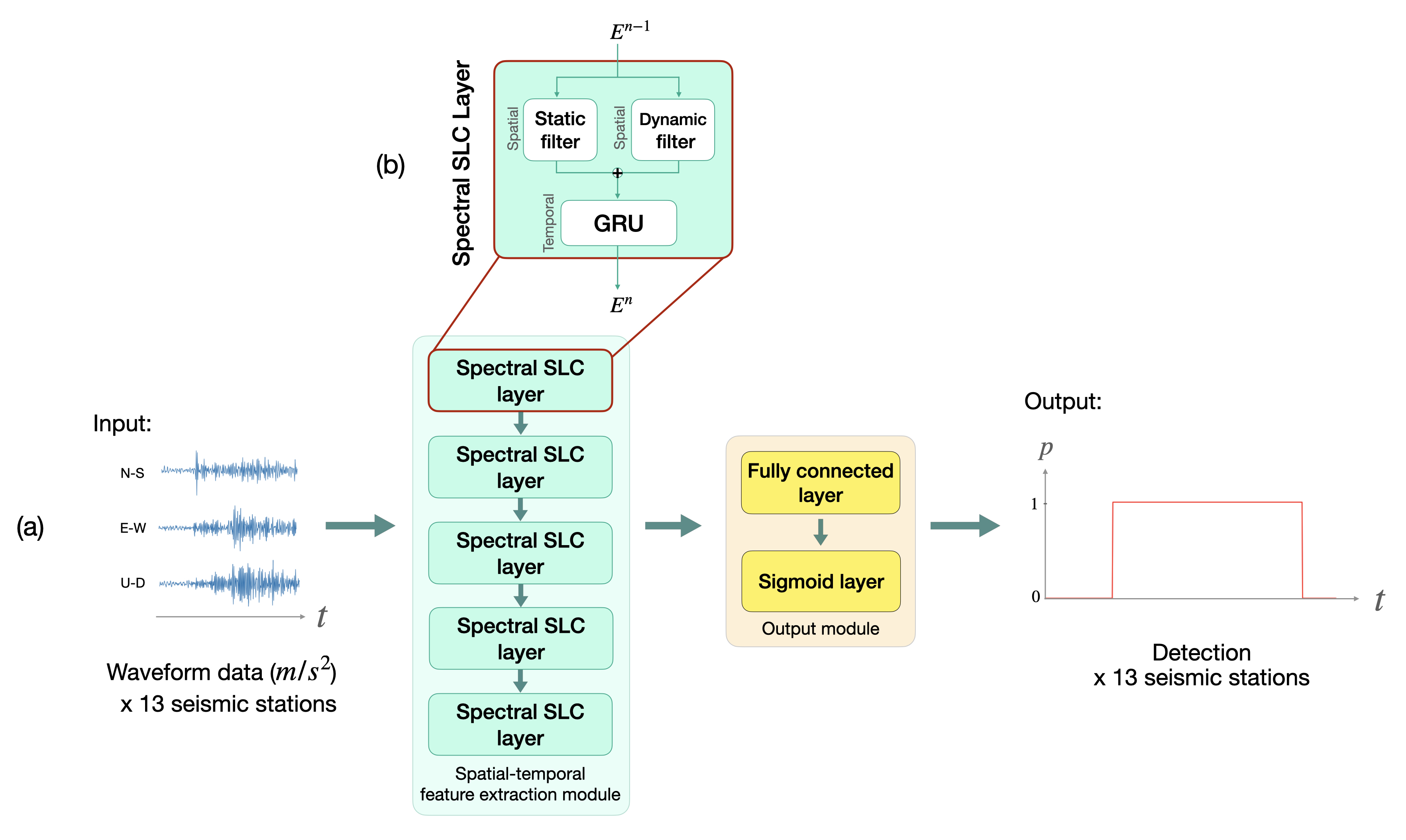} 
    \caption{(a) The architecture of our GCN. The input comprises three-component waveform data from each of the 13 seismic stations, while the output is a detection probability time series for each station. The GCN consists of a spatial-temporal feature extraction module and an output module. The spatial-temporal feature extraction module contains 5 Spectral SLC layers. The output module includes a fully connected layer followed by a sigmoid layer. (b) The Spectral SLC layer integrates a spatial component to capture static and dynamic graph structures, and a temporal component to model temporal dependencies in the signals.}
    \label{fig:arc}
\end{figure*}

To account for the temporal dynamics of the waveform data, the features extracted by the spatial static and dynamic filters in Eq. \ref{eqn:spatial-output} are passed through a recurrent layer, specifically a Gated Recurrent Unit (GRU) \cite{cho-etal-2014-properties}. The GRU processes the sequence of features for each station over time, capturing temporal dependencies for earthquake detection, resulting in the final output of a Spectral SLC layer, which is shown in Fig. \ref{fig:arc}b. 

We selected the GRU over alternative temporal modeling approaches, such as Long Short-Term Memory (LSTM) \cite{10.1162/neco.1997.9.8.1735} or Pseudo three Dimensional convolution (P3D) \cite{qiu2017learning}, due to its reduced parameter count. This makes GRUs more robust against overfitting, particularly in scenarios with limited earthquake waveform data, while still effectively modeling temporal dynamics.

\subsection{Network Architecture}
The overall architecture is illustrated in Fig. \ref{fig:arc}, comprising multiple Spectral SLC layers. To reduce overfitting and improve generalization, a dropout layer with a rate of 0.2 is applied after each Spectral SLC layer. The output tensor is then passed through a fully connected layer, followed by a Sigmoid activation function, to produce the final output: probability time-series indicating the probability of an earthquake event at each seismic station.

We train our GCN using the Adam optimizer \cite{diederik2014adam}, with the objective of minimizing the binary cross-entropy loss between the predicted probability time series and the ground truth labels. To optimize the model's performance, we employed \emph{Optuna} \cite{10.1145/3292500.3330701} to tune several hyper-parameters, including the number of SLC layers, the hidden dimension of intermediate feature representations, the maximum Chebyshev polynomial order, the learning rate, and the batch size. Hyper-parameter tuning was conducted using 5-fold cross-validation on the training dataset.

\section{Result and Discussion}
In this study, we use a conventional GCN as a baseline model, which employs a predefined static adjacency matrix representing a fully connected graph with all nodes equally connected. The GCN aggregates spatial features using a fixed graph structure and models temporal dependencies using a Gated Recurrent Unit (GRU). In contrast, our proposed method learns both static and dynamic (input-driven) relationships between seismic stations.

We refer to the ``minimum detection probability (MDP)'' as the threshold above which the model's output probability classifies the signal at a time step as an earthquake. A true positive occurs when the model correctly classifies an earthquake signal at a time step, while a false positive occurs when it incorrectly classifies noise as an earthquake signal at a time step.

\begin{figure}
    \centering
    \includegraphics[width=.45\textwidth]{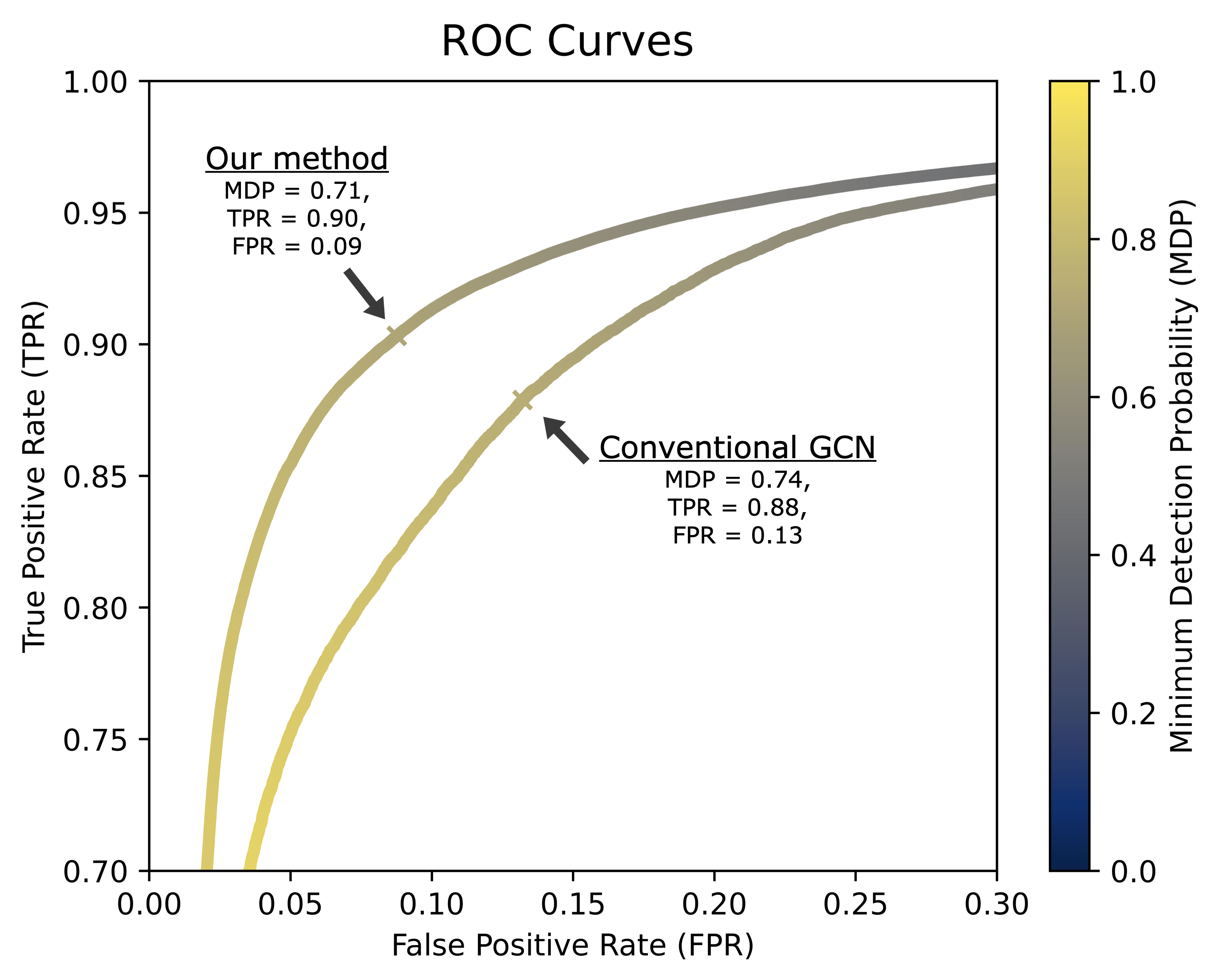} 
    \caption{ROC curves of our proposed method and the baseline (Conventional GCN) model, evaluated on the test dataset. The ROC curves illustrate the trade-off between the true positive rate (TPR) and false positive rate (FPR) at various minimum detection probabilities (MDP). The color bar represents the corresponding MDP values. Our method demonstrates superior performance, achieving a higher TPR and lower FPR. The optimal MDP, representing the FPR-TPR point nearest to the upper-left corner, is indicated by an arrow for each model.}
    \label{fig:result-roc}
\end{figure}

To evaluate the performance of our model in detecting earthquake signals, we analyzed its Receiver-operating characteristic (ROC) curve. The ROC curve provides a graphical representation of the trade-off between the true positive rate (TPR) and false positive rate (FPR) across various detection thresholds. The ROC curves were generated using the detection probabilities produced by our model and the baseline's on the test dataset. Fig. \ref{fig:result-roc} shows the ROC curves for our model and baseline (Conventional GCN) model. Our model demonstrates superior performance as its ROC curve reaches closer to the top-left corner of the plot compared to the conventional GCN. This indicates that our model achieves a higher true positive rate (TPR) at a given false positive rate (FPR), reflecting its ability to detect earthquake signals more accurately while minimizing false alarms. The steeper rise and the shift of the curve towards the optimal region highlight the enhanced capability of our approach in distinguishing earthquake signals from noise. For an example, the performance of our method by three different minimum detection probabilities (MDP: 0.55, 0.6, 0.71) is shown in Table \ref{tab:results}. The model is able to detect earthquake signals with these three threshold probabilities, however the false positive rate increases as the threshold is set lower.

\begin{figure*}
    \centering
    \includegraphics[width=.7\textwidth]{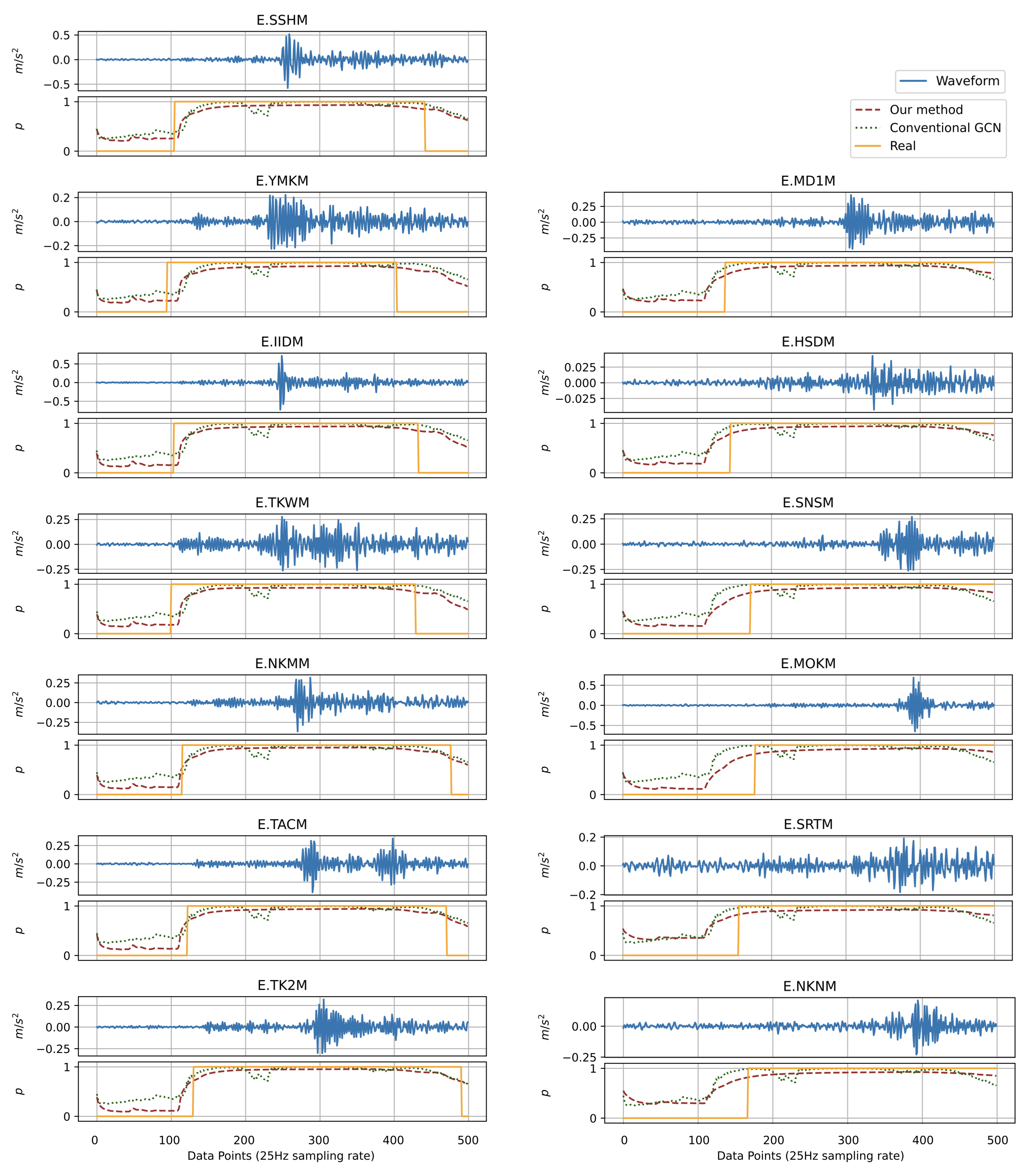} 
    \caption{Detection results for 13 seismic stations using unseen waveform data from the North-South component (September 4, 2011, 00:20:14 to 00:20:34, Japan time). The top panel of each subplot shows the waveform (blue), while the bottom panel compares detection probabilities from our method (dashed brown), the baseline model (dotted green), and the ground-truth labels (solid orange).}
    \label{fig:result-ex}
\end{figure*}

\begin{table}[ht]
\centering
\caption{Resultant True Positives and False Positives of Our Method for Three Different MDPs (Minimum Detection Probabilities).}
\label{tab:results}
\begin{tabular}{@{}lccc@{}}
\toprule
\textbf{MDP} & \textbf{0.55}         & \textbf{0.6}        & \textbf{0.71}         \\ \midrule
\textbf{True positive rate (TPR)} & 0.95 & 0.94 & 0.90 \\
\textbf{False positive rate (FPR)} & 0.20 & 0.16 & 0.09   \\ \bottomrule
\end{tabular}
\vspace{0.5em}
\caption*{\textit{Notes.} A true positive occurs when the model correctly classifies an earthquake signal at a time step, while a false positive occurs when it incorrectly classifies noise as an earthquake signal at a time step.}
\end{table}
An example of the results from unseen waveform data as input is shown in Fig. \ref{fig:result-ex}. The blue lines represent the waveform data of 13 seismic stations in the North-South component, recorded over a 20-second time window from  September 4, 2011, 00:20:14 (Japan time) at a 25 Hz sampling rate. Below each waveform, the detection results of our model (dashed brown) and the baseline model (dotted green) and the ground-truth labels (solid orange) are displayed to demonstrate how our proposed method and the baseline model perform relative to the ground-truth labels. Notably, the detection probabilities output by our method more consistently align with the ground-truth labels with less fluctuations compared to the baseline.

Our results demonstrate that the proposed Spectral GCN with structural learning significantly outperforms the conventional GCN baseline in terms of true positive rate (TPR) and false positive rate (FPR). By incorporating multi-station data and dynamically modeling both static and input-driven relationships between seismic stations, our approach enhances the detection of earthquake signals while minimizing false alarms. Additionally, the ability to generate detection probability time series for each station provides finer granularity, making the method particularly suitable for real-time applications in earthquake early warning (EEW) systems. Future work includes applying our model to longer waveform data using sliding windows for real-time detection.

Our future work will aim to make the method more adaptive to changing nature of data by  replacing GRU with adaptive graph signal processing methods \cite{10.1016/j.sigpro.2022.108662}.  We will also consider modeling seismological noise\cite{kuruoglu1997new, 1201785} and apply more eficient joint filtering and detection \cite{costagli2007image}.

\bibliographystyle{IEEEtran}  
\bibliography{ref}  

\vspace{12pt}

\end{document}